%% file: root.tex
\documentclass[a4paper]{svproc}
\usepackage{graphicx}%
\usepackage{multirow}%
\usepackage{amsmath,amssymb,amsfonts}%
\usepackage{mathrsfs}%
\usepackage{xcolor}%
\usepackage{textcomp}%
\usepackage{manyfoot}%
\usepackage{booktabs}%
\usepackage{algorithm}%
\usepackage{algorithmicx}%
\usepackage{algpseudocode}%
\usepackage{listings}%
\usepackage{xurl}
\usepackage{bm}
\usepackage{wrapfig}
\begin{document}
\mainmatter 
\title{Versatile Airborne Ultrasonic NDT Technologies via Active Omni-Sliding with Over-Actuated Aerial Vehicles}

\titlerunning{Versatile Airborne Ultrasonic NDT Technologies}


\author{Tong Hui\inst{1,2\thanks{This work has been supported by the European Unions Horizon 2020
Research and Innovation Programme AERO-TRAIN under Grant Agreement
No. 953454.}} \and Florian Braun \inst{2} \and
Nicolas Scheidt\inst{2} \and
Marius Fehr \inst{2} \and Matteo Fumagalli\inst{1}}
\institute{Technical University of Denmark, \email{tonhu@dtu.dk, mafum@dtu.dk}  \and
Voliro AG, Zurich, Swizterland, \email{name.surname@voliro.com}}

\authorrunning{T. Hui et al.}

\maketitle              
\begin{abstract}
This paper presents the utilization of advanced methodologies in aerial manipulation to address meaningful industrial applications and develop versatile ultrasonic Non-Destructive Testing (NDT) technologies with aerial robots. The primary objectives of this work are to enable multi-point measurements through sliding without re-approaching the work surface, and facilitate the representation of material thickness with B and C scans via dynamic scanning in arbitrary directions (i.e. omnidirections). To accomplish these objectives, a payload that can slide in omnidirections (here we call the omni-sliding payload) is designed for an over-actuated aerial vehicle, ensuring truly omnidirectional sliding mobility while exerting consistent forces in contact with a flat work surface. The omni-sliding payload is equipped with an omniwheel-based active end-effector and an Electro Magnetic Acoustic Transducer (EMAT).  Furthermore, to ensure successful development of the designed payload and integration with the aerial vehicle, a comprehensive studying on contact conditions and system dynamics during active sliding is presented, and the derived system constraints  are later used as guidelines for the hardware development and control setting. The proposed methods are validated through experiments, encompassing both the wall-sliding task and dynamic scanning for Ultrasonic Testing (UT), employing the aerial platform - Voliro T.
\keywords{UAV, Aerial Robots, Design and Prototyping.}
\end{abstract}

\input{01_introduction}

\input{02_approach}

\input{03_system_constraints}
\input{04_experiments}

\input{05_conclusion}

\input{06_bib}

%
%
%
%




\end{document}

%% file: 01_introduction.tex
\section{Introduction}\label{sec1}
Recently unmanned aerial manipulator (UAM) systems have gained traction in the field of ultrasonic non-destructive testing (NDT) techniques for industrial inspection \cite{par2021}\cite{watson2022}\cite{trujillo2019}. Significant investigation in control and platform development on aerial manipulation has been dedicated in the past years \cite{anibal2021}. However, the practical utilization of advanced methodologies for meaningful inspection data gathering that align with industrial needs remains limited \cite{watson2022}. This limitation has motivated the present work, which aims to develop versatile airborne ultrasonic NDT technologies. In ultrasonic testing (UT), obtaining a healthy amplitude scan (A-scan) from an accurate point measurement is crucial for evaluating the thickness of the testing material \cite{ut_scan}. Moreover, conducting data collection at multiple contact points offers a broader range of information regarding the testing material. And the capability of gathering continuous measurements at different locations on the testing surface, known as dynamic scanning, allows the representation of the testing material thickness with two-dimensional scan (B-scan) and three-dimensional scan (C-scan), which are considered as more sophisticated methods for viewing UT data \cite{ut_scan}.

In \cite{trujillo2019}, a fully actuated aerial vehicle was integrated with a passive end-effector (EE) equipped with wheels and an ultrasonic gauge sensor for UT operations with the capability of dynamic scanning. The gauge sensor used in \cite{trujillo2019} requires a couplant (e.g. gel) to fill in the gap between the sensor and the testing surface for measuring, and it is often difficult to ensure this condition while sliding along the surface for dynamic scanning. In \cite{watson2022}, a dry and non-contact ultrasonic sensor - the Electro Magnetic Acoustic Transducer (EMAT) is mounted on passive EEs for UT with an over-actuated aerial vehicle. In both \cite{watson2022}\cite{trujillo2019}, the precision of the contact point highly relies on the position information of the aerial vehicle and can be crucial in practical industrial testing environments. In \cite{layout2022}\cite{wop2018}, instead, active EEs with omnidirectional mobility (omni-mobility) were employed for high-precision contact-based tasks and sliding operations. The integration of an active EE with omni-mobility enables the sliding motion along the work surface to reach arbitrary points without being constrained by EE orientation. However, neither \cite{watson2022}\cite{trujillo2019} nor \cite{wop2018} can have truly omnidirectional sliding mobility without changing aerial vehicle orientation due to lacking of omni-mobility at the EE with the current wheel type \cite{watson2022}\cite{trujillo2019} or underactuation of the aerial vehicle \cite{wop2018}. The platform developed in \cite{layout2022} offers truly omni-sliding ability, but is not implemented for real industrial applications. And it is equipped with major compliant components to have flexibility during sliding which increase difficulties in modeling due to non-rigidity. Moreover, to the best of the authors' knowledge, there exists a noticeable gap in the comprehensive studying of the force conditions during active sliding and their effects on the aerial robot system considering inherent constraints such as actuator saturation and contact conditions. This study is essential to guarantee successful deployment of aerial robots in sliding operations.

To address the limitations identified in the aforementioned related works, this research aims to develop an aerial manipulator that possesses truly omni-sliding capability while maintaining stable physical interaction with a flat work surface for versatile airborne ultrasonic NDT applications. Gaining inspiration from previous studies in \cite{watson2022}\cite{layout2022}\cite{wop2018}, an omnidirectional sliding (omni-sliding) payload is designed. This payload is equipped with an active EE using omni wheels and an EMAT sensor. Moreover, a compliant component design with limited compliance level is introduced which does not overly complicate system modeling. Furthermore, with the designed omni-sliding payload, system constraints introduced by contact conditions and the aerial platform are presented via a comprehensive studying of the force conditions during sliding. The identified constraints are then used to ensure successful hardware development of the payload and integration with an over-actuated aerial vehicle enabling dynamic ultrasonic scanning in arbitrary directions. Experimental validation is conducted to verify the sliding capability of the aerial manipulator with the developed payload and to demonstrate the effectiveness of dynamic scanning with data measurements.

%% file: 02_approach.tex
\section{Technical Approach}\label{sec2}
In this section, we present the over-actuated aerial vehicle employed for aerial manipulation and the design of the omni-sliding payload.\vspace{-10pt}
\subsection{Aerial Platform Description}
\label{platform}
\begin{figure}[h]%
\centering
\begin{minipage}{0.50\linewidth}
    \includegraphics[width=\linewidth]{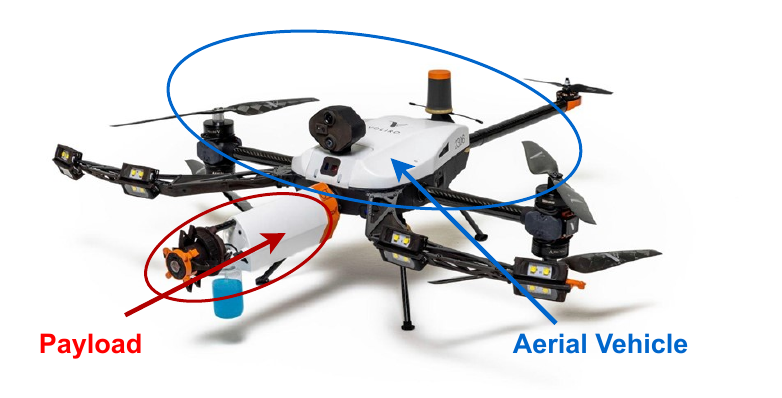}
\caption{Voliro T: an over-actuated aerial vehicle equipped with a payload.}
\label{voliro}
\end{minipage} \hfill
\begin{minipage}{0.35\linewidth}
    \includegraphics[width=\linewidth]{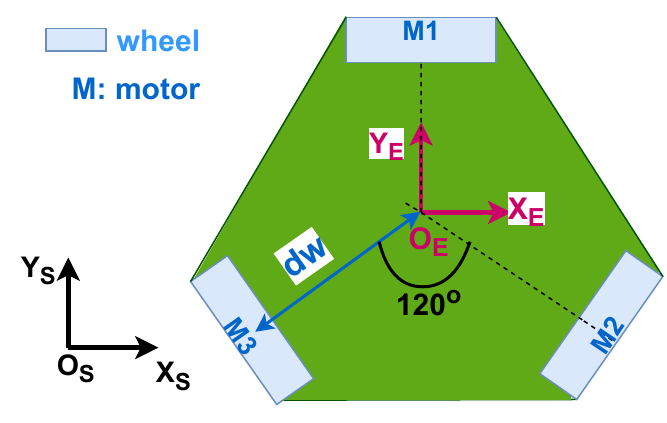}
\caption{Wheel (motor) layout, work surface frame $\mathcal{F}_S$, end-effector frame $\mathcal{F}_E$.}
\label{fig:layout}
\end{minipage}
\end{figure}

The aerial platform chosen for this study is the Voliro T \cite{voliro}, see Fig.~\ref{voliro}. The Voliro T is an over-actuated aerial vehicle equipped with a modular payload system and it enables the generation of omnidirectional wrenches in any orientation of the aerial vehicle, with decoupled translation and rotation dynamics. The modular design of the payload system provides flexibility and adaptability, allowing efficient integration with various end-effectors and sensors for different applications. The current flight controller ensures stable physical contact with the work surface and consistent linear force generation. Voliro T can exert up to 30 N of linear force while supporting the total system mass of 4 kg \cite{watson2022} \cite{voliro}. And it can provide torque disturbance rejection of up to 2 N·m in any direction, alongside its ability to generate linear forces during interactions. Moreover, to ensure a stable contact with the work surface, it has been determined from previous experience that a minimum force of 10 N is exerted from the aerial vehicle in the direction normal to the surface.
\vspace{-20pt}
\subsection{Omni-Sliding Payload Design}\label{sec:design}
\subsubsection{Active End-Effector}
For the design of the omni-sliding payload, it is assumed that the work surface is flat and provides sufficient area for sliding operations. We define a frame attached to a reference point on the work surface as $\mathcal{F}_S=\{\bm{O}_S;\bm{X}_S,\bm{Y}_S\}$. To achieve omni-mobility, an active EE design is utilized, consisting of three omniwheels. These wheels are positioned at an angle of $120$ degrees between each pair, ensuring uniform distribution around the center of geometry (COG) of the EE tip with a distance of $d_W$, see Fig.~\ref{fig:layout}. An EE frame $\mathcal{F}_E=\{\bm{O}_E;\bm{X}_E,\bm{Y}_E\}$ is attached to the COG of the EE tip. We define $\bm{\upsilon}_E=[\upsilon_x \ \upsilon_y]^{\top}$ as the linear velocity of the EE frame with respect to the work surface frame $\mathcal{F}_S$, while the orientation of the EE is set to zero. The motor speeds driving the omniwheels are then determined by $\bm{\upsilon}_E$ and geometric properties of the EE via kinematic transformation as below:  
\begin{align}\label{eq:motor}
    u_1&=\frac{\upsilon_x}{R_w}, \nonumber \\
    u_2&=- \frac{\upsilon_x}{2R_w}- \frac{\sqrt{3}\upsilon_y}{2R_w},\\
    u_3&=- \frac{\upsilon_x}{2R_w}+ \frac{\sqrt{3}\upsilon_y}{2R_w}, \nonumber 
\end{align}
where $R_w$ is the wheel radius, and $u_i$ with $i=1,2,3$ are motor inputs, for details please refer to \cite{robotics}. 

\subsubsection{Compliant Component} \label{sec:compliance}
\begin{wrapfigure}{r}{0.5\linewidth}%
\centering \vspace{-20pt}
    \includegraphics[width=\linewidth]{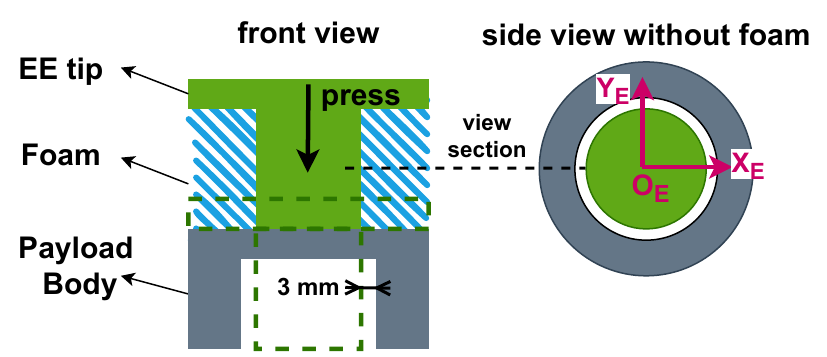}
\caption{Compliant Component: front view and side view.}
\vspace{-20pt}
\label{fig:compliance}
\end{wrapfigure}

While Voliro T is in stable contact, the baseline interaction controller prioritizes the attitude control, gravity compensation and force generation to the surface. To avoid overly complicating system modeling while allowing some flexibility between the active EE and the aerial vehicle, a compliant component with limited compliance level is incorporated between the EE tip and the payload body as in Fig.~\ref{fig:compliance}, knowing that the payload body is rigidly attached to the aerial vehicle. 
This compliant component is conducted by an outer hollow cylinder that rigidly connects to the payload body and an inner cylinder that rigidly connects to the EE tip with 3~mm tolerance in between two cylinders. The tolerance between two cylinders allows movements of the EE tip in any direction parallel to $(\bm{X}_E,\bm{Y}_E)$ up to 3 mm with respect to the payload body.  
The two cylinders are connected by a foam and can have piston-like movements along the EE symmetric axis when the EE tip is in contact to reduce the impact effects.
With the above design, the relative motion between the EE tip and the payload body and the non-rigidity of the system are limited while allowing some compliance. During active sliding operations, the contact forces applied from the work surface generate torques around the aerial vehicle center of mass (CoM) which are considered as disturbances to the attitude dynamics of the aerial vehicle and are mitigated by the attitude control within the range of disturbance rejection. We assume that when the active EE moves with a small velocity and acceleration, the attitude controller is robust to preserve the aerial vehicle orientation under the uncertainties due to sliding, and sufficiently responsive to prevent the wheel detachment from the work surface using the designed compliant component. The detailed studying regarding the system dynamics during active sliding is presented in Sec.~\ref{sec:dynamics}.
\vspace{-10pt}
\subsubsection{UT Sensor}
The EMAT sensor \cite{watson2022}\cite{voliro} is mounted at the center of the EE tip and allows a maximum lift-off distance of 4 mm from the testing surface. The EMAT sensor does not require a couplant, which can then prevent any undesirable interactions between the sensor and the work surface during sliding (the magnetic force between EMAT and the work surface is neglected considering the lift-off).
To successfully develop the designed omni-sliding payload integrating with the aerial vehicle using the current baseline controller, a comprehensive studying regarding to the system constraints during sliding is essential.
\vspace{-10pt}

%% file: 03_system_constraints.tex
\section{System Constraints for Active Sliding}\label{sec:dynamics}
In this section, with the designed payload and the studied aerial vehicle in Sec.~\ref{sec2}, we present the system constraints derived from the required contact conditions and the aerial platform limitations to ensure the desired active sliding capability via a comprehensive studying on the linear and angular dynamics of the system during active sliding. We define the body frame $\{B\}=\{\bm{O}_B;\bm{X}_B,\bm{Y}_B,\bm{Z}_B\}$ being attached to the CoM of the aerial vehicle\footnote{All vectors in this section are expressed in the body frame.}, in which $(\bm{Y}_B,\bm{Z}_B)$ has the same orientation as $(\bm{X}_E,\bm{Y}_E)$. Let $\{I\}=\{\bm{O};\bm{X},\bm{Y},\bm{Z}\}$ present the inertial frame.
\subsection{Contact Conditions}\label{sec:contact}
\begin{figure}[ht]%
\centering
\begin{minipage}{0.48\linewidth}
    \includegraphics[width=\linewidth]{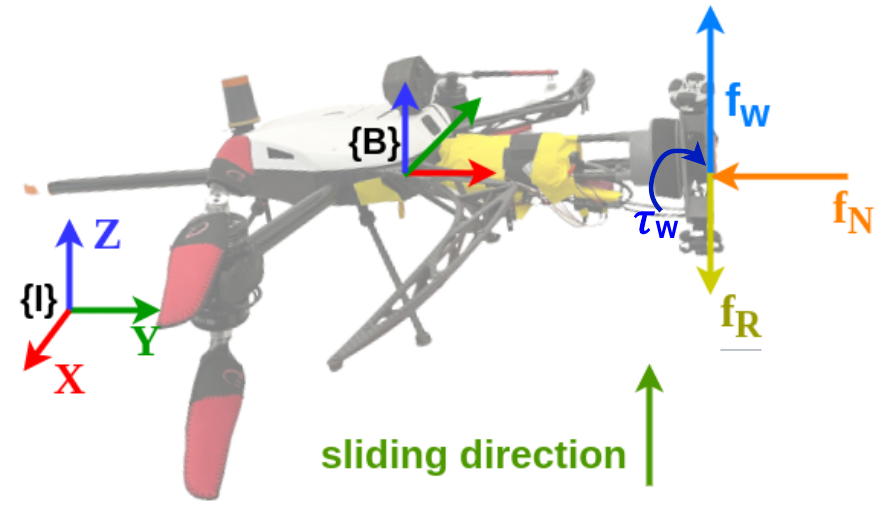}
\caption{Active Sliding, \{B\}: body frame, \{I\}: inertial frame, $\bm{f}_N$: normal force, $\bm{f}_R$: rolling friction, $\bm{f}_w$: friction force to drive the wheels, $\bm{\tau}_w$: total wheel motor torque.}
\label{ee_force}
\end{minipage} \hfill
\begin{minipage}{0.45\linewidth}
    \includegraphics[width=\linewidth]{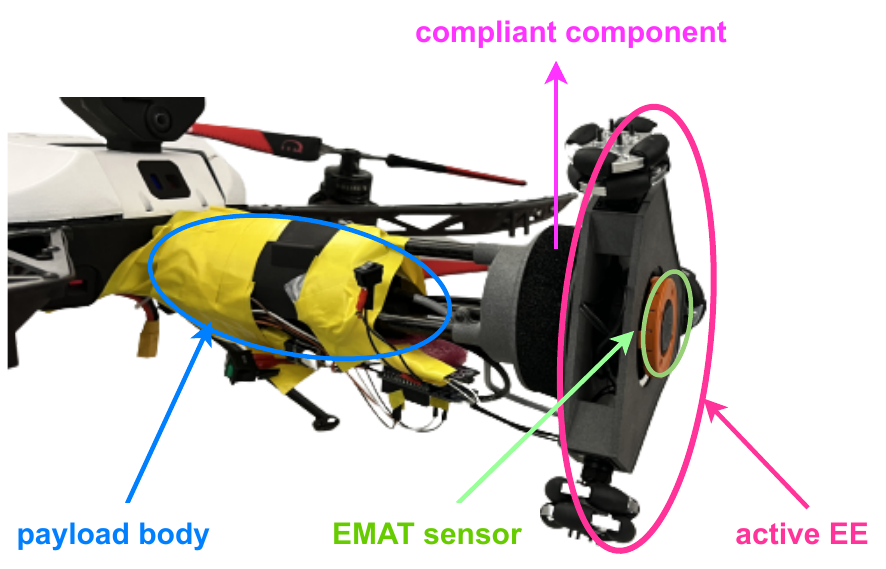}
\caption{Omni-Sliding Payload Prototype.}
\label{ee}
\end{minipage}
\end{figure}

We assume that all three wheels are always in contact with the surface during sliding considering small sliding velocities and accelerations, and the contact wrenches generated from the environment to the EE are pure linear forces acting at the COG of the EE tip as in Fig.~\ref{ee_force}. 
$\bm{f}_N$ is the total force vector normal to the surface acting on the EE. $\bm{f}_R$ is the rolling resistance force with $|\bm{f}_R|=\mu_R \cdot |\bm{f}_N|$ and $\mu_R$ is the rolling resistance coefficient \cite{friction}. The total driving force $\bm{f}_w$ is the friction force (aligning with the moving direction) resulted from the torque $\bm{\tau}_w$ generated by wheel motors with $|\bm{f}_w|=\frac{|\bm{\tau}_w|}{R_w}$, and $\bm{f}_w$ always has the opposite sign of $\bm{f}_R$ for active wheels. Assuming that the selected omniwheels can ensure: the total wheel load $C_{wheel} [kg] \cdot 9.81 [m/s^2] \gg |\bm{f}_N|[N]$,
in order to drive the EE with active omniwheels on the working surface without slipping, the following contact conditions are required:
\begin{itemize}
    \item (a) the static friction coefficient $\mu_S$ between the wheel and the surface must be significantly larger than the rolling resistance coefficient $\mu_R$, i.e. $\mu_R\ll \mu_S$;
    \item (b) $\mu_R|\bm{f}_N|<|\bm{f}_w|=\frac{|\bm{\tau}_w|}{R_w} \leq  min(\eta\frac{\tau_w^{max}}{R_w},\mu_S|\bm{f}_N|)$, where $\tau_w^{max}$ is the maximum torque value generated by 3 wheel motors at required rotating speed, $0<\eta<1$ is the safety factor. 
\end{itemize}
\subsection{System Modeling}
 We assume that the active EE is driven with small velocities and accelerations, and the deformation of the compliant component parallel to $(\bm{X}_E,\bm{Y}_E)$ can be neglected considering its small value (up to 3~mm) based on the design in Sec.~\ref{sec:compliance}. The dynamics of wheels and wheel motors are neglected considering the light weight and small inertia. Therefore, with the aforementioned contact conditions and assumptions, the overall system equations of motion during active sliding expressed in the body frame $\{B\}$ can be written as:
\begin{equation}\label{system}
    \bm{M}\dot{\bm{\upsilon}}+\bm{h}(\bm{\upsilon},\dot{\bm{\upsilon}})+\bm{g}=\begin{bmatrix}
        \bm{T}_a\\ \bm{\tau}_a
\end{bmatrix}+ 
\begin{bmatrix} \bm{F}_C\\ \bm{F}_C \times \bm{l}_C \end{bmatrix}+\begin{bmatrix}
    \bm{0}_3 \\ \bm{\tau}_w
\end{bmatrix},
\end{equation}
where $\bm{M}=diag\big(\begin{bmatrix}\bm{M}_{lin} &\bm{M}_{ang}\end{bmatrix}\big) \in R^{6\times6}$ is the general inertia matrix\footnote{lin: linear component, ang: angular component.}, $\bm{h}=\begin{bmatrix}\bm{h}_{lin} \\\bm{h}_{ang}\end{bmatrix} \in R^{6 \times 1}$ counts for Coriolis and centrifugal effects, and $\bm{g}=\begin{bmatrix}\bm{g}_{lin}\\\bm{g}_{ang}\end{bmatrix}\in R^{6 \times 1}$ counts for the gravity term. $\bm{\upsilon}=\begin{bmatrix}\bm{\upsilon}_{lin}\\\bm{\upsilon}_{ang}\end{bmatrix} \in R^{6 \times1}$ presents the generalized velocity of the system during sliding, where $\bm{\upsilon}_{lin}=\begin{bmatrix}0 \\ \bm{\upsilon}_E\end{bmatrix} \in R^{3 \times 1}$ and we assume that $\bm{\upsilon}_{ang}=\bm{0}_3$ during sliding. $\begin{bmatrix}
        \bm{T}_a\\ \bm{\tau}_a
\end{bmatrix} \in R^{6 \times 1}$ is the stacked thrust and torque vector generated by the actuation system of the aerial vehicle. $\bm{l}_C \in R^{3 \times 1}$ is the vector from the COG of the EE tip to the CoM of the aerial vehicle and the contact force vector $\bm{F}_C=\bm{f}_w+\bm{f}_R+\bm{f}_N \in R^{3 \times 1}$ at the EE tip is the only source of external wrenches acting on the system. With $\bm{\tau}_w$ being always opposite to $\bm{f}_w \times \bm{l}_C$ by definition and $\bm{f}_N$ passes through the CoM of the aerial vehicle without generating torque, the Eq.(\ref{system}) can be rewritten as:
\begin{equation}\label{system_new}
    \bm{M}\dot{\bm{\upsilon}}+\bm{h}(\bm{\upsilon},\dot{\bm{\upsilon}})+\bm{g}=\begin{bmatrix}
        \bm{T}_a\\ \bm{\tau}_a
\end{bmatrix}+ 
\begin{bmatrix} \bm{f}_w+\bm{f}_R+\bm{f}_N\\ \bm{f}_w 
\times \bm{l}_w +\bm{f}_R\times\bm{l}_C \end{bmatrix},
\end{equation}
where $\bm{l}_w=\frac{\bm{l}_C}{|\bm{l}_C|}(|\bm{l}_C|-R_w)$.
\subsection{Platform Limitations}\label{sec:platform}
Knowing the maximum force generation and torque disturbance rejection of the studied aerial vehicle as $f_{max} \in R$ and $\Gamma_{max} \in R$, in this section we look into the effects of the contact forces on the system linear and angular dynamics during active sliding and the constraints introduced by the platform via decomposing the Eq.(\ref{system_new}) into linear and angular dynamics. 
We consider now the linear dynamics during sliding. While the thrust $\bm{T}_a$ mainly supplies the force generation to the work surface $\bm{f}_N$ and gravity compensation regarding to $\bm{g}_{lin}$, we can write:
\begin{equation}\label{eq:lin}
\bm{M}_{lin}\dot{\bm{\upsilon}}_{lin}+\bm{h}_{lin}=\bm{f}_w+\bm{f}_R.
\end{equation}
To ensure the contact condition (b) in Sec.~\ref{sec:contact}, with Eq.(\ref{eq:lin}) we introduce the constraint:
$|\bm{M}_{lin}\dot{\bm{\upsilon}}_{lin}+\bm{h}_{lin}-\bm{f}_R| \leq  min(\eta\frac{\tau_w^{max}}{R_w},\mu_S|\bm{f}_N|).$
Moreover the main effects on the aerial vehicle thrust during active sliding comes from the applied normal force vector on the work surface $\bm{f}_N$ and the total weight of the whole system. Considering the maximum force generation of the aerial vehicle, a constraint is set by:
$|\bm{f}_N| \leq f_{max}$.

Considering now the angular dynamics of the system while sliding, with $\bm{\upsilon}_{ang}=\bm{0}_3$ we can write:
\begin{equation}\label{eq:ang}
    \bm{g}_{ang}=\bm{\tau}_a+\bm{f}_w 
\times \bm{l}_w +\bm{f}_R\times\bm{l}_C.
\end{equation}
Considering the small value of $|\bm{f}_R||R_w|$ compared to $|\bm{f}_w| 
|\bm{l}_w|$, we have $|\bm{f}_R\times \bm{l}_C| \approx |\bm{f}_R\times \bm{l}_w|$ and by substituting Eq.(\ref{eq:lin}) into Eq.(\ref{eq:ang}), we can write: $\bm{\tau}_a:=\bm{g}_{ang}-(\bm{M}_{lin}\dot{\bm{\upsilon}}_{lin}+\bm{h}_{lin})\times \bm{l}_w$. 
Therefore, the uncertainties in angular dynamics of the aerial vehicle are mainly affected by the velocity and acceleration during sliding. The constraint introduced by the torque disturbance rejection of the aerial vehicle is defined as:
$\big(|\bm{M}_{lin}\dot{\bm{\upsilon}}_{lin}|+|\bm{h}_{lin}|\big) \cdot |\bm{l}_w| \leq \Gamma_{max}$.
\subsection{System Constraints}
With the above study, we then can summarize the system constraints that satisfy the required contact conditions while considering platform limitations to ensure successful active sliding as:
\begin{equation}\label{eq:cons}
\begin{cases}
\text{(a)} \ \mu_R\ll \mu_S,\\\text{(b)} \
|\bm{M}_{lin}\dot{\bm{\upsilon}}_{lin}+\bm{h}_{lin}-\bm{f}_R| \leq  min(\eta\frac{\tau_w^{max}}{R_w},\mu_S|\bm{f}_N|),\\\text{(c)} \
|\bm{f}_N| \leq f_{max},\\ \text{(d)} \
\big(|\bm{M}_{lin}\dot{\bm{\upsilon}}_{lin}|+|\bm{h}_{lin}|\big) \cdot |\bm{l}_w| \leq \Gamma_{max}.
\end{cases}
\end{equation}
These constraints are then used as guidelines for hardware development of the active omni-sliding payload considering the materials at contact, the maximum sliding velocity and acceleration, and also the normal force vector to be applied on the work surface. These constraints introduce an upper boundary $f_{up}$ of $|\bm{f}_N|$ once the maximum sliding velocity and acceleration are defined. A minimum pushing force magnitude is required to ensure stable contact of the wheels with the work surface which introduces a lower boundary $f_{low}$ of $|\bm{f}_N|$. The developed aerial manipulator system must allow a non-empty interval $f_{low}\leq |\bm{f}_N| \leq f_{up}$ to accomplish successful development of the payload and integration for the active sliding task.


%% file: 04_experiments.tex
\section{Experiments and Results}\label{sec3}
In this section, the identified system constraints in Eq.(\ref{eq:cons}) are considered to develop the prototype of the designed omni-sliding payload and the integration with the introduced aerial vehicle. The developed aerial manipulator is then tested with a vertical wall for validating its omni-sliding capability. Furthermore, a real case NDT operation is shown by dynamic scanning on a testing surface with the UT sensor measurements.
\subsection{Experiment Setup}\label{subsec2}
\subsubsection{Prototype of the omni-sliding payload}
Based on the design in Sec.~\ref{sec:design} and system constraints for active-sliding in Sec.~\ref{sec:dynamics}, the prototype of the designed omni-sliding payload is developed, see Fig.~\ref{ee}. To ensure the contact constraints, omniwheels with approximately $\mu_R:0.02 - 0.04$ and $\mu_S:0.35 - 0.5$ for the testing surfaces are selected to ensure Eq.(\ref{eq:cons}.a). Each wheel can bear up to $3$ kg of load (i.e. $C_{wheel}=3 \times 3 =9$ kg) with $R_w=0.024$ m. The wheel motors used have a maximum stall torque of 0.5 N$\cdot$m each. The maximum sliding velocity and acceleration is set to $0.01$ m/s and $0.02$ m/s$^2$ respectively guaranteeing Eq.(\ref{eq:cons}.d) considering the Voliro T in Sec~\ref{platform}. The velocity vector $\bm{\upsilon}_E$ is controlled using a remote controller (RC). The 2-dimensional pulse-width modulation (PWM) signals generated by the RC sticks are mapped into the range of velocity value $[0,0.01]$, which then is used to calculate the desired motor inputs using Eq.(\ref{eq:motor}) via the micro-controller. \vspace{-10pt}
\subsubsection{Aerial Vehicle Setting}
With the above setting, the developed payload adds roughly $0.4$ kg on the system mass compared with other standard payloads. To mathematically identify the range of $|\bm{f}_N|$ in Eq.~\ref{eq:cons}.b is difficult due to the uncertain wheel motor behaviour in terms of torque generation under different rotating speed. Therefore, a load test is executed to obtain the maximum $|\bm{f}_N|$ that ensures the constraint. During the load test, different weights were added on the stand alone omni-sliding payload prototype while trying to drive it on various horizontal-placed surfaces. From the load test, $|\bm{f}_N|$ up to 15 N is allowed by the hardware setting. Considering the Voliro T with the additional weights of the sliding payload, a safe range of applicable $|\bm{f}_N|$ is up to 25 N to ensure Eq.(\ref{eq:cons}.c) and minimum 10 N to preserve good contact. Therefore to satisfy both the contact conditions and aerial vehicle limits ensuring stable contact, the feasible range of $|\bm{f}_N|$ is $[10, 15]$ N, which is a non-empty interval indicating that the developed aerial manipulator system satisfies the constraints for the active sliding operation. Finally the reference $|\bm{f}_N|$ is set to $10$ N with a purpose of energy saving. The developed system is then used to execute the wall sliding test and the dynamic scanning test with NDT sensor.

\subsection{Test 1: Wall Sliding}
\begin{figure}[ht]%
\centering
\vspace{-15pt}
\begin{minipage}{0.46\linewidth}
    \includegraphics[width=0.9\linewidth,height=4.5cm]{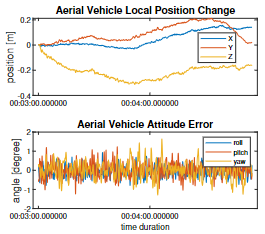}
\caption{wall sliding: aerial vehicle local position change and attitude error during sliding.}
\label{attitude}
\end{minipage} \hfill
\begin{minipage}{0.48\linewidth}
\includegraphics[width=\linewidth,height=4cm]{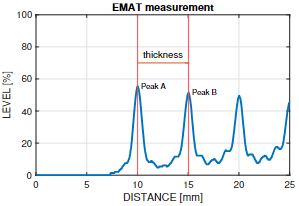}
\caption{EMAT measurement for A scan, testing material thickness is the distance between two peaks A and B.}
\label{emat}
\end{minipage}\vspace{-20pt}
\end{figure}

The first test is to drive the developed omni-sliding payload with the RC in arbitrary direction on a flat vertical wall surface. During the sliding operation, the aerial vehicle moves along with the active EE thanks to the design of the omni-sliding payload. The attitude errors of the aerial vehicle while sliding are shown in Fig.\ref{attitude} for a time duration with a maximum absolute value of around 2 degree.
\vspace{-10pt}
\subsection{Test 2: Dynamic Ultrasonic Scanning with EMAT}
During this test, a flat sample metal piece is used for ultrasonic testing. The EMAT sensor is capable of providing an A-scan at a single contact point with the data processing technology from Voliro T \cite{watson2022}, as in Fig.~\ref{emat}, where the thickness of the testing material can be calculated by the travel distance of the sound between Peak A and Peak B. By sliding along the testing surface, multiple point measurements at different locations can be collected without detaching from the surface and re-approaching. Moreover, the continuous measurements via omnidirectional dynamic scanning enables the data representation with B-scan and C-scan (the representation of B and C scans are not of focus in this paper), and the time-varying EMAT measurements along with sliding are shown in the attached video.

%% file: 05_conclusion.tex
\section{Main Experimental Insights}\label{sec5}
The developed omni-sliding payload demonstrates the capability of achieving omni-sliding mobility for an over-actuated aerial vehicle, particularly showcasing dynamic scanning for UT. However, the actuation of the EE introduces challenges such as increased weight and bulkiness to the payload design. To address weight-related challenges, careful selection of lightweight wheels with high load-bearing capacity and favorable friction characteristics, as well as employing lightweight motors with intelligent control and sufficient torque allowance, becomes essential to ensure successful development. Furthermore, the contact forces during the interaction with the work surface play a significant role in the physical demands placed on the developed aerial manipulator system. Understanding the force and torque generation limits of the aerial vehicle, along with determining the minimum force required for stable contact, becomes imperative for the hardware development of the omni-sliding payload and its successful integration with the aerial vehicle to execute the required tasks effectively. In a word, gaining a comprehensive understanding of the system constraints during active sliding, encompassing the aerial vehicle and the contact conditions, holds significant relevance in validating the proposed technologies through practical experiments and ultrasonic applications.


%% file: 06_bib.tex
\bibliographystyle{splncs04}